\documentclass[]{spie}  

\usepackage{amsmath,amsfonts,amssymb}
\usepackage{graphicx}
\usepackage{gensymb}
\usepackage{bbm}
\usepackage{multirow}
\usepackage{graphicx}

\usepackage[colorlinks=true, allcolors=blue]{hyperref}
\usepackage{setspace}
\usepackage{times}

\usepackage{array}

\newcolumntype{"}{@{\tabcolsep\vrule width 2pt\hskip\tabcolsep}}
\newcolumntype{P}[1]{>{\centering\arraybackslash}p{#1}}
\makeatother

\title{SSL$^2$: Self-Supervised Learning meets Semi-Supervised Learning: Multiple Sclerosis Segmentation in 7T-MRI from large-scale 3T-MRI}

\author[a]{Jiacheng Wang}
\author[b]{Hao Li}
\author[a]{Han Liu}
\author[b]{Dewei Hu}
\author[a]{Daiwei Lu}
\author[c]{Keejin Yoon}
\author[c]{Kelsey Barter}
\author[c]{Francesca Bagnato}
\author[a]{Ipek Oguz}
\affil[a]{Dept. of Computer Science, Vanderbilt University, Nashville, TN, US}
\affil[b]{Dept. of Electrical and Computer Engineering, Vanderbilt University, Nashville, TN, US}
\affil[c]{Dept. of Neurology, Vanderbilt University Medical Center, Nashville, TN, US}

\begin{document} 
\maketitle

\begin{abstract}
Automated segmentation of multiple sclerosis (MS) lesions from MRI scans is important to quantify disease progression. In recent years, convolutional neural networks (CNNs) have shown top performance for this task when a large amount of labeled data is available. However, the accuracy of CNNs suffers when dealing with few and/or sparsely labeled datasets.
A potential solution is to leverage the information available in large public datasets in conjunction with a target dataset which only has limited labeled data. In this paper, we propose a training framework, SSL$^2$ (self-supervised-semi-supervised), for multi-modality MS lesion segmentation with limited supervision. We adopt self-supervised learning to leverage the knowledge from large public 3T datasets to tackle the limitations of a small 7T target dataset. To leverage the information from unlabeled 7T data, we also evaluate state-of-the-art semi-supervised methods for other limited annotation settings, such as small labeled training size and sparse annotations. We use the shifted-window (Swin) transformer\cite{liu2021swin} as our backbone network.
The effectiveness of self-supervised and semi-supervised training strategies is evaluated in our in-house 7T MRI dataset. The results indicate that each strategy improves lesion segmentation for both limited training data size and for sparse labeling scenarios. The combined overall framework further improves the performance substantially compared to either of its components alone. Our proposed framework thus provides a promising solution for future data/label-hungry 7T MS studies.

\end{abstract}


\section{INTRODUCTION}
\label{sec:intro}  

Multiple Sclerosis (MS) is a prevalent inflammatory disease of the central nervous system that poses significant diagnostic and monitoring challenges. A commonly utilized tool for addressing these challenges is magnetic resonance imaging (MRI), which enables the visualization and quantification of focal MS lesions. Segmentation of these lesions is a crucial step for clinical evaluation. However, manual segmentation is time-consuming and labor-intensive, mainly due to large variations in location, shape, and intensity among lesions.

7 Tesla (7T) MRI scans provide more detailed anatomical data with higher spatial resolution and allow more precise lesion quantification. While many fully-supervised lesion segmentation algorithms \cite{roy2018multiple,danelakis2018survey,zhang2019multiple,zhang2021segmentation,liu2022moddrop++} exist, these are often impractical for use with 7T data, given the limited availability of publicly available annotated MS lesion datasets at 7T and the high cost of annotating in-house 7T MRI datasets.

In order to overcome the limitations posed by the scarcity of annotated data and the labor-intensive nature of manual annotation, recent research has focused on exploring self-supervised and semi-supervised learning techniques. Self-supervised learning \cite{taleb20203d} aims to extract robust high-dimensional features directly from the MRI data using augmentations and proxy tasks. Augmentations involve transforming the data and using these transformations as supervision to learn features. Proxy tasks, such as predicting the rotation angle of an image, can then be used to train the model to learn useful features. On the other hand, semi-supervised learning leverages the abundance of unlabeled data to improve model performance \cite{cheplygina2019not}. This is achieved through two main approaches: consistency learning \cite{tarvainen2017mean} and self-training \cite{chen2021semi}. Consistency learning maximizes the stability of predictions of an unlabeled image and its noise-perturbed counterparts. Self-training generates pseudo labels from unlabeled data and retrains the model using these weak supervision signals. By utilizing both labeled and unlabeled data, semi-supervised learning offers a promising solution to the challenge of limited annotated data.


In recent years, researchers have proposed the use of transformers in the medical imaging domain to capture long-term dependencies \cite{hatamizadeh2022unetr,li2022cats}. To further improve the efficiency of these approaches, the Shifted windows (Swin)-Transformer\cite{liu2021swin, hatamizadeh2022swin} utilizes a non-overlapping shifted window schema to significantly reduce the computational complexity for several vision tasks. The utilization of a Transformer-based architecture has the potential to improve the performance of MS lesion segmentation tasks by effectively capturing the long-range dependencies present in the data.

In this paper, we propose a novel approach to address the challenges of limited data and time-consuming annotations in MS lesion segmentation using MRI scans. Our approach utilizes self-supervised and semi-supervised learning methods to leverage the large number of publicly available 3T MRI scans to obtain a robust pre-trained model, using a Swin-Transformer as our backbone. We fine-tune this pre-trained model for our 7T in-house MR data on the downstream lesion segmentation task. Our hypothesis is that our pre-trained model from 3T images can boost the performance of MS lesion segmentation on a sparsely labeled 7T dataset. We consider two sparse labeling scenarios: a small number of fully labeled volumes, or volumes with only a small number of labeled slices. The contributions of our proposed self-supervised-semi-supervised learning (SSL$^2$) framework are:
\begin{itemize}
    \item We demonstrate a significant increase in MS lesion segmentation performance compared to traditional supervised training (Dice score improvement from \textbf{$0.6971$} to \textbf{$0.8186$}) on the in-house 7T MRI dataset.
    \item Our proposed model generates robust segmentation results, even when using very few training samples or sparsely annotated datasets, and outperforms other methods in these limited supervision settings.
    \item We effectively distill knowledge from large public 3T MRI datasets to facilitate 7T MRI studies, and make the pre-trained model weights publicly available for further research \footnote{\url{https://github.com/MedICL-VU/SSL-squared-7T}}.
\end{itemize}

\section{Materials and Methods}
\label{sec:methods}

In this section, we describe our proposed framework for addressing the challenge of limited annotation in 7T MRI scan dataset segmentation. We begin by introducing the datasets used in our framework in Section \ref{sec:dataset}. Next, in Section \ref{sec:ssl2}, we provide an overview of the entire framework. We then delve into the key components of our framework in Sections \ref{sec:semi-supervise} and \ref{sec:self-supervise}, where we discuss the semi-supervised segmentation and self-supervised training, respectively. Finally, we describe our implementation details in Section \ref{sec:imple-detail}.

\subsection{Datasets}
\label{sec:dataset}
\subsubsection{In-house 7T MRI dataset}
We obtained an in-house dataset of 7T MRI scans at the Vanderbilt University Medical Center (VUMC) for 37 MS patients at various disease stages, all at their first visit. Table \ref{tab:demo} presents the demographic and clinical information. Each subject's data includes MP2RAGE (T1-weighted) and FLAIR scans.

\begin{table}[h]
 \caption{Demographic and clinical information for the VUMC 7T MS dataset.}
    \centering
    \begin{tabular}{p{9cm} p{3.8cm}}
    \hline
         Number of Subjects & 37 \\
    \hline
        Female & 17 ($45.9\%$) \\
        Male & 20 ($54.1\%$) \\
        \hline
        Black/African American & 3 ($8.1\%$) \\
        White/Caucasian & 34 ($91.9\%$)\\
        \hline
        RRMS (relapsing-remitting MS) & 29 ($78.4\%$)\\
        CIS (clinically isolated syndrome) & 4 ($10.8\%$)\\
        RIS (radiologically isolated syndrome) & 2 ($5.4\%$)\\
        Others & 2 ($5.4\%$)\\
        \hline
        Age (mean $\pm$ STD) & $37.46\pm9.07$\\
        EDSS (expanded disability status scale, mean $\pm$ STD) & $1.42 \pm 1.56$\\
        \hline
    \end{tabular}
   
    \label{tab:demo}
\end{table}

\textbf{Labeled 7T in-house dataset.} Expert manual annotation of MS lesions on T1-weighted images was performed by a single trained annotator (KY) for a subset of 14 subjects. The automated lesion segmentation on these same 14 subjects was then performed using our existing model, ModDrop++ \cite{liu2022moddrop++}. ModDrop++ is based on our previous Tiramisu-based model \cite{zhang2019multiple} which currently holds the top position on the ISBI 2015 challenge \cite{carass2017longitudinal} leaderboard (\url{https://smart-stats-tools.org/lesion-challenge}). Discrepancies between the manual and automated lesion masks were reviewed and reconciled by a second trained radiologist (KB) to generate the final lesion masks for the 14 subjects.

\textbf{Unlabeled 7T in-house dataset.} The remaining 23 subjects are used as our unlabeled dataset. 

\subsubsection{Public unlabeled 3T MRI datasets}
We employ several publicly available datasets without their annotations for pre-training, including:
\begin{itemize}
    \item Longitudinal MS Lesion Segmentation Challenge (ISBI 2015) dataset 
    \cite{roy2018multiple}, with 21 training and 61 testing scans; 
    \item UMCL (University Medical Center Ljubljana) Multi-rater Consensus dataset \cite{lesjak2018novel}, with 30 scans; 
    \item MICCAI MSSeg 2016 Challenge \cite{commowick2016msseg} dataset, with 15 scans;
    \item MICCAI Brain Tumor Segmentation (BraTS) 2021 Challenge \cite{menze2014multimodal} dataset with 1251 scans.
\end{itemize} 

For consistency, only T1-weighted and FLAIR images are utilized from each of these datasets. Rigid registration is performed to align each FLAIR image to its corresponding T1w image; all T1w images are rigidly aligned to a single subject from our 7T dataset. 

\subsubsection{Preprocessing}
We apply standard preprocessing techniques to all images (3T and 7T), including model-based skull stripping with HD-BET\cite{isensee2019automated}, N4 bias correction \cite{tustison2010n4itk}, and histogram normalization to the range $[0,1]$. The resulting images have a resolution of $0.9\times0.9\times 0.9 \mathrm{mm}^3$ and a field of view of $256\times256\times200$ voxels. 

\subsubsection{Data representation}
For both the 3T and 7T MRI datasets, we crop the image to the bounding box of the skullstripped brain to reduce image size. To further address the memory constraints commonly encountered when training 3D networks, we randomly extract two sub-volumes of $96\times 96 \times 96$ voxels from each input image.  The T1w and FLAIR images are concatenated as input to our network. 

\subsection{System Overview}
\label{sec:ssl2}

\begin{figure} [t]
  \begin{center}
  \begin{tabular}{c} 
  \includegraphics[scale=0.6]{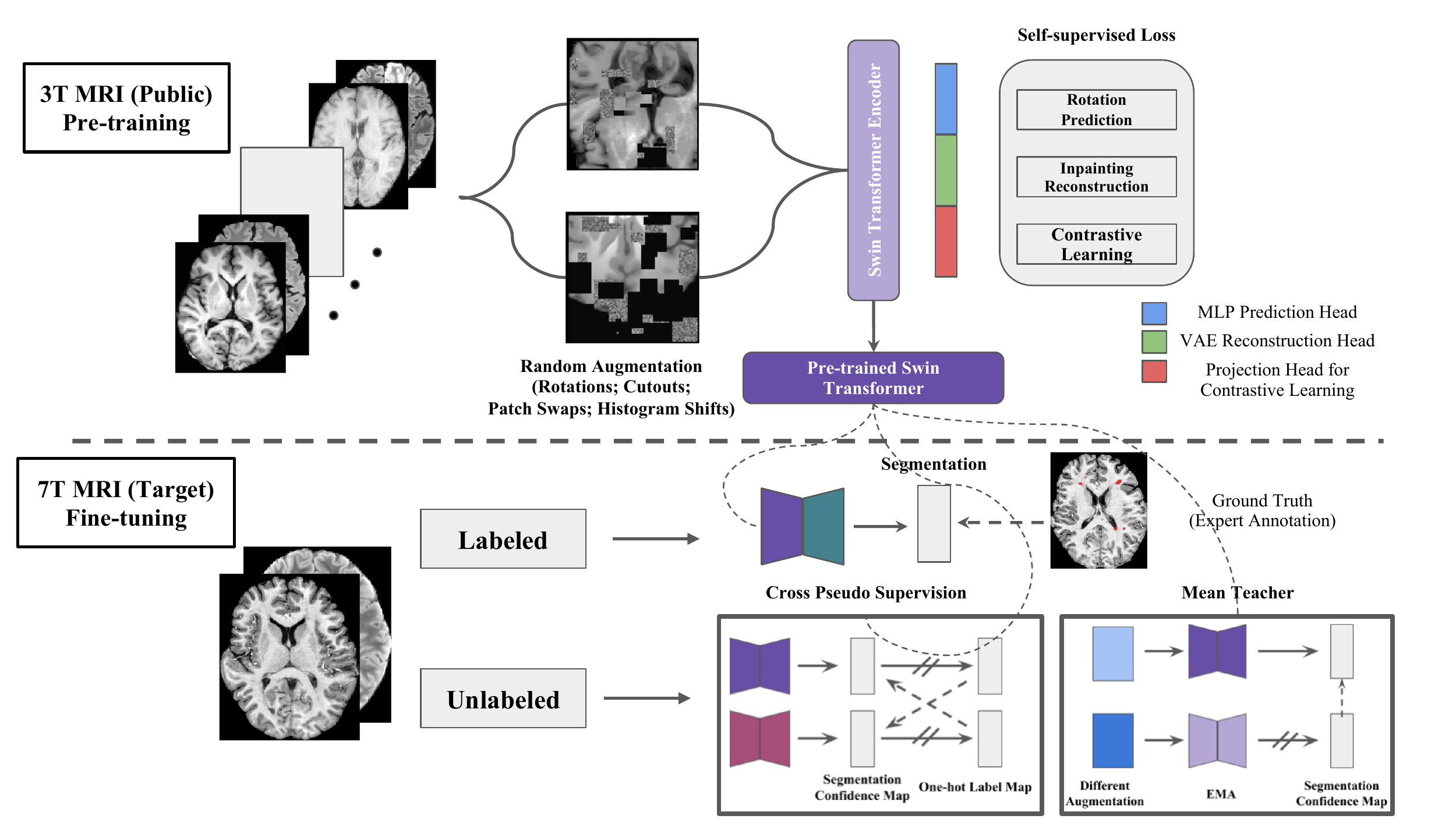}
  \end{tabular}
  \end{center}
  \caption
  { \label{fig:overall-workflow} 
Overall framework. \textbf{Top.} Self-supervised pre-training.  3 heads are attached to the encoder for the 3 proxy tasks. \textbf{Bottom.} Semi-supervised fine-tuning. Pre-trained weights (purple)  are used to initialize the encoder. Mean-teacher and CPS are compared.}
  \end{figure}

Figure \ref{fig:overall-workflow} shows the overall workflow of our system.  

We use the 7T MRI dataset (labeled and unlabeled) in a semi-supervised manner to determine the best semi-supervision strategy for the MS lesion segmentation task. We compare 6 different semi-supervised learning schemes in a 7-fold cross-validation setting. This experiment identifies the best and worst performing semi-supervision strategies, which are incorporated into our $SSL^2$ framework.

$SSL^2$ begins by using the unlabeled 3T MRI datasets to pre-train a Swin transformer model in a self-supervised manner using 3 proxy heads that correspond to 3 proxy tasks. For this purpose, we use a modification of the original 2D Swin-transformer block \cite{hatamizadeh2022swin} to be suitable for 3D volumes as our backbone network, similar to Swin-UNETR \cite{hatamizadeh2022swin}. Next, the proxy heads are discarded and only the pre-trained Swin transformer weights are preserved. The 7T MRI dataset (labeled and unlabeled) and the previously determined best-performing semi-supervision strategy are incorporated to obtain the final segmentation results. We compare this model to the baseline model which uses the worst-performing semi-supervision strategy.

\subsection{Semi-supervised Segmentation in 7T MRI dataset}
\label{sec:semi-supervise}
To choose the semi-supervision strategy for our framework, we compare the performance of six different semi-supervised learning schemes in a 7-fold cross-validation setting. These methods include Mean Teacher \cite{tarvainen2017mean}, Entropy Minimization\cite{vu2019advent}, Deep Adversarial Networks \cite{zhang2017deep}, Uncertainty Aware Mean Teacher \cite{yu2019uncertainty}, FixMatch \cite{sohn2020fixmatch, media2022urpc}, and Cross Pseudo Supervision \cite{chen2021semi}. 

Each semi-supervised model is trained with a combination of supervised loss $L_{sup}$ and unsupervised loss $L_{unsup}$ such that $L_{semi}=L_{sup}+\lambda_{semi}L_{unsup}$. The supervised loss is computed using Dice Loss and pixel-wise Cross Entropy (CE) Loss, while the semi-supervised loss varies among the models. During each training iteration, equal number of labeled and unlabeled samples are used to compute $L_{sup}$ and $L_{unsup}$, respectively. 

Our experiments utilize the 14 subjects from our in-house 7T labeled dataset. We utilize 12 subjects for training and hold out 2 subjects for validation in each fold. Additionally, we make use of the 23 subjects in our unlabeled in-house 7T dataset for the semi-supervised models. We use a sliding-window inference method as implemented in MONAI. The window size is chosen as $96 \times 96 \times 96$ voxels with overlap of $24 \times 24 \times 24$ voxels. We use the Dice score to evaluate the performance of the compared methods.

The results are presented in Table \ref{tab:semi}.
We observe that while the performance of these methods is highly comparable, the Mean Teacher \cite{tarvainen2017mean} performs the worst and Cross Pseudo Supervision (CPS) \cite{chen2021semi} performs the best. We choose these two methods as the baseline and top performer, respectively, for the rest of our experiments. These two methods are described in more detail below.

\begin{table}[b]
    \centering
    \caption{Segmentation performance (Dice score) for 6 different semi-supervised models in our 7-fold cross-validation experiment (Sec.\ \ref{sec:semi-supervise}). The fully supervised model does not have access to any unlabeled samples and thus has the lowest performance. Bold indicates the best performance in each fold. Among the semi-supervised methods, Mean Teacher is the worst performer and Cross Pseudo Supervision is the top performer for the MS lesion segmentation task. }
    \resizebox{\columnwidth}{!}{
    \begin{tabular}{l|l|l|l|l|l|l|l||l|l}
    \hline
        Methods & Fold 1 & Fold 2 & Fold 3 & Fold 4 & Fold 5 & Fold 6 & Fold 7 & Avg  & Std \\ \hline
        Fully Supervised \cite{liu2022moddrop++} & 0.6982 & 0.6844 & 0.6867 & 0.6989 & 0.6744 & 0.6904 & 0.6964 & 0.6899 & 0.0089 \\ 
        Mean Teacher \cite{tarvainen2017mean} & 0.7107 & 0.7048 & 0.6991 & 0.7055 & 0.6990 & 0.7088 & 0.7048 & 0.7047 & 0.0044 \\ 
        Entropy Minimization \cite{vu2019advent} & 0.7238 & 0.7318 & 0.7312 & 0.7422 & 0.7320 & 0.7391 & 0.7257 & 0.7323 & 0.0066 \\
        Deep Adversarial Networks \cite{zhang2021segmentation}& 0.7283 & 0.7455 & 0.7294 & 0.7521 & 0.7252 & 0.7263 & 0.7241 & 0.7330 & 0.0111 \\ 
        Uncertainty Aware Mean Teacher \cite{tarvainen2017mean}& 0.7347 & 0.7246 & 0.7279 & 0.7351 & 0.7263 & 0.7352 & 0.7321 & 0.7308 & 0.0045 \\ 
        FixMatch \cite{sohn2020fixmatch} & 0.7732 & 0.7753 & 0.7614 & \textbf{0.7897} & 0.7726 & 0.7656 & 0.7797 & 0.7739 & 0.0092 \\
        Cross Pseudo Supervision \cite{chen2021semi} & \textbf{0.7808} & \textbf{0.7894} & \textbf{0.7764} & 0.7834 & \textbf{0.7901} & \textbf{0.7803 }& \textbf{0.7844} & \textbf{0.7835} & 0.0050 \\ 
        \hline
    \end{tabular}
    }

        \label{tab:semi}

\end{table}

\begin{figure}[t]
    \centering
    \includegraphics[width=\columnwidth]{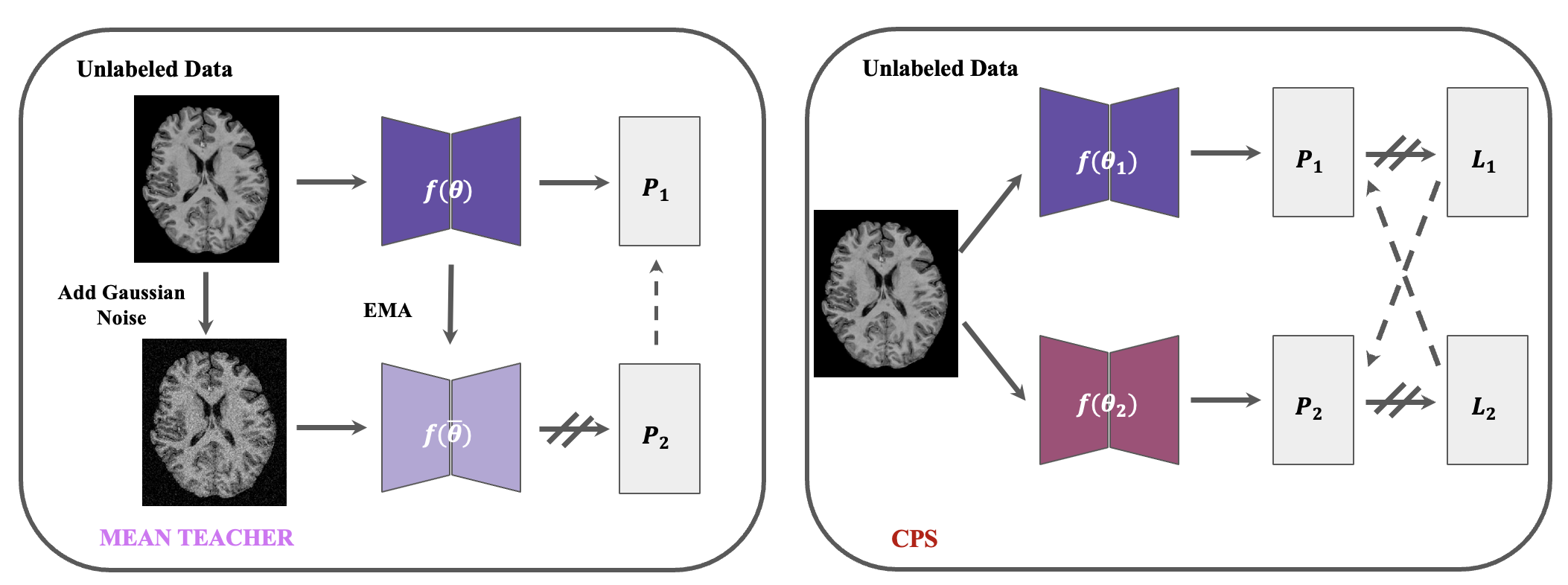}
    \caption{Illustration of the semi-supervised methods schema. The left panel shows the Mean Teacher model. The right panel  shows the CPS model. '$\rightarrow$' means forward operations, '$\dasharrow$' means loss supervision, and '$//$' means stop-gradient. }
    \label{fig:semi_struct}
\end{figure}

\textbf{Baseline: Mean Teacher\cite{tarvainen2017mean}.} The Mean Teacher method utilizes two identical networks, denoted as student model $f(\theta)$ and teacher model $f(\hat{\theta})$, as shown in the left panel of Figure \ref{fig:semi_struct}. The core idea of this method is to use the input with perturbed Gaussian noise as another input to compute the consistency loss defined as $L_{unsup} = ||P_2 - P_1||^2$. The network parameters $f(\theta)$ are updated using $L_{semi}= L_{sup} + \lambda_{semi} L_{unsup}$ where $\lambda_{semi}$ is the trade-off weight. The weights for the teacher model $f(\bar{\theta})$ are updated using the exponential moving average (EMA) to temporally ensemble the versions of the student models $f(\theta)$ from previous iterations. This strategy enforces stable predictions without the help of annotations.

\textbf{Top Performer: Cross Pseudo Supervision\cite{chen2021semi} (CPS).} The CPS combines the ideas of self-training using pseudo labels and cross-probability consistency. It utilizes two networks, denoted $f(\theta_1)$ and $f(\theta_2)$, with different dropouts, as shown in the right panel of Figure \ref{fig:semi_struct}. The two networks produce the probability outputs $P_1$ and $P_2$, respectively. We apply argmax to these probability outputs to obtain one-hot labels, $L_1$ and $L_2$. These labels are then used to supervise the training of the other branch, i.e., $L_1$ supervises $P_2$, and $L_2$  supervises $P_1$.
 The semi-supervised loss is calculated using $L_{unsup_1} = CE(L_2, P_1)$ and $L_{unsup_2} = CE(L_1, P_2)$.  The  network parameters $f(\theta_1)$ and $f(\theta_2)$ are updated using $L_{semi_1} = L_{sup_1} + \lambda_{semi} L_{unsup_1}$ and $L_{semi_2} = L_{sup_2} + \lambda_{semi} L_{unsup_2}$ respectively. We note that $L_{sup_1}$ and $L_{sup_2}$ are computed by passing labeled images into the networks $f(\theta_1)$ and $f(\theta_2)$, respectively.

\subsection{Self-supervised training model in public 3T MRI datasets}
\label{sec:self-supervise}


 In this section, we describe our self-supervised augmentation scheme for leveraging unlabeled 3T MRI datasets. We introduce random augmentations to our input images and use the pairs of real and augmented images for 3 proxy tasks to allow self-supervision, inspired by Tang et al.~\cite{tang2022self}. To accomplish this, we concatenate three different task-specific heads after the encoder and compute three self-supervised losses: rotation prediction, inpainting reconstruction, and contrastive loss. 
 The total self-supervised loss for the pre-training is  defined as $L_{pre\_training} = \lambda_1 L_{rot} + \lambda_2 L_{inpaint} + \lambda_3 L_{contrast}$ where $\lambda_i$ are hyper-parameters. We choose $\lambda_1 = \lambda_2 = \lambda_3 =1$ in our experiments.

 \subsubsection{Augmentation methods}

\begin{figure}[t]
    \centering
    \includegraphics[scale=0.8]{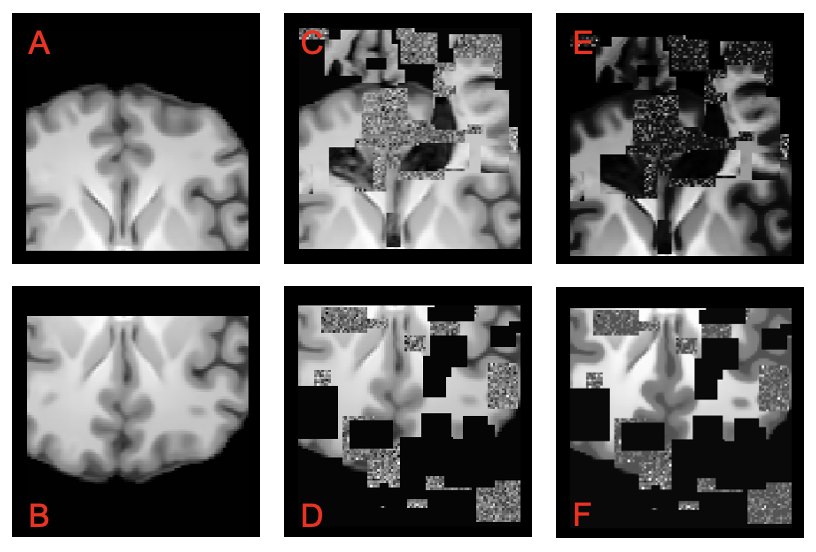}
    \caption{Augmentation for a  subject from the UMCL dataset \textbf{A.} The cropped sub-volume of $96 \times 96 \times 96$. $\mathbf{A \rightarrow B}.$ Rotation of $r=180 \degree$. $\mathbf{A \rightarrow C, B \rightarrow D}.$ Random cutout and patch swaps.
    $\mathbf{C \rightarrow E, D \rightarrow F}.$ Random histogram shifts.}
    \label{fig:augment}
\end{figure}

 Table \ref{tab:augmentation} lists the three types of augmentation we apply and the associated parameters for each. 
 \begin{enumerate}
     \item \textbf{Random rotations.} We follow the method of Chen et al.~\cite{chen2019self} and apply a combination of random rotations of $r \in \{0\degree, 90\degree, 180\degree, 270\degree \}$. The case $r=180\degree$ is illustrated in Figure  \ref{fig:augment} between panels $A \rightarrow B$.
     \item \textbf{Random cutouts and patch swaps.} We choose random rectangular patches and replace it with either a different random patch in the brain or with random noise. Recall that the data is represented as sub-volumes of $96 \times 96 \times 96$ voxels in our models. The patch sizes are constrained to [0.05, 0.25] of the sub-volume size along each dimension, and a total volume less than $30\%$ of the sub-volume size. This augmentation is illustrated in Figure \ref{fig:augment} between panels  $A \rightarrow C$ and $B \rightarrow D$. 
     \item \textbf{Random histogram shifts.} We use the random histogram shift as implemented in MONAI (\url{https://monai.io}). This is illustrated in Figure \ref{fig:augment} between panels  $C \rightarrow E$ and $D \rightarrow F$.  Note that we do not have a proxy task that leverages this augmentation; instead, the goal of this augmentation is to reduce the domain gap between 3T and 7T datasets by increasing the diversity of the training dataset.
 \end{enumerate}

\begin{table}[b]
\caption{Data augmentation during our training process. $\mathbf{p}$ indicates the probability of applying an augmentation.}
    \centering
    \begin{tabular}{  p{4.5cm}p{5cm}p{3cm} }
         \hline
          \textbf{Transform} & \textbf{Parameters} & $\mathbf{p}$ \\
         \hline
         Random Rotate  & $r \in \{0\degree, 90\degree, 180\degree, 270\degree \}$ & $0.25,0.25,0.25,0.25$ \\
         Random Crop and Patch Swap & $v$=0.3, $b_{max}$ = 0.25, $b_{min}$ = 0.05 & 1 \\
         Random Histogram Shift &  \# control points = $1, 3$ & $0.5,0.5$ \\
         \hline
    \end{tabular}
    
    \label{tab:augmentation}
\end{table}

\subsubsection{Proxy Task 1: Rotation Prediction} 
The rotation prediction task is designed to ensure that the encoder learns to extract robust features that are invariant to rotation. We train the encoder to predict the rotation angle categories $r \in \{0\degree, 90\degree, 180\degree, 270\degree \}$ in the rotated sub-volume. A single multilayer perceptron (MLP) head is attached to the encoder to predict the softmax rotation angle possibilities $\hat{P}_r$ given the ground truth $P_r$ and cross-entropy loss is used for updating the parameters.
\begin{align*}
    L_{rot} = - \sum_{r}^{0\degree, 90\degree, 180\degree, 270\degree} P_r \log(\hat{P_r})
\end{align*}

\subsubsection{Proxy Task 2: Inpainting Reconstruction} 

The inpainting reconstruction task is designed to improve the encoder's generalization ability and semantic understanding. We train the encoder to reconstruct the missing information from the random cutout and patch swaps. For the inpainting reconstruction, we apply a Variational Autoencoder (VAE)\cite{an2015variational} head containing $6$ 3D convolution blocks with instance normalization. We add a leaky-ReLU activation on the downsampling path for each scale and an MLP layer on the upsampling path. Given the cutout patch $I_{pred}$ and the original image $I_{org}$, we use the L1 Loss to train the reconstruction network. 
\begin{align*}
    L_{inpaint} = || I_{pred} - I_{org}||_1
\end{align*}

\subsubsection{Proxy Task 3: Contrastive Representation Learning} 

The contrastive representation learning task is designed to encourage the encoder to learn representations that are robust to different data augmentations and reduce the domain gap between different datasets. We create a minibatch of $2N$ samples by  applying two random augmentations (combination of rotate/crop/histogram shift) to each of $N$ subjects, such that the minibatch contains two views of each subject. Then, we randomly select two images from the minibatch and train the encoder to predict whether they are from the same subject. Specifically, a linear MLP projection head is applied to map the latent features from the encoder into higher dimensions $v$. We use cosine-similarity \cite{chen2020improved} to maximize the agreement among positive pairs (same subject $i$, different augmentations) $v_{i,1}, v_{i,2}$ and minimize the negative pairs (different subjects $i$ vs.\ $k$). Thus the loss function is defined as 
\begin{align*}
    L_{contrast} = -\log \frac{\exp (sim(v_{i,1}, v_{i,2})/t)}{\sum_{k=1, k \neq i }^{N} \sum_{j=1}^{2}\exp(sim (v_{i,1}, v_{k,j})/t)}, 
\end{align*}
where $t$ is the temperature and $sim(\cdot)$ denotes cosine similarity.



\subsection{Implementation Details}
\label{sec:imple-detail}
Given the input sub-volumes of size $96 \times 96 \times 96$, we use the Swin-Transformer window and patch size of $2 \times 2 \times 2$, which leads to $48\times 48\times 48$ patches.  We use 4 down-sampling blocks and set the number of features to 12, resulting in a latent feature size of $(3\times 3\times 3)\times 2^4 \times 12$. We set the number of transformer heads as $[3, 6, 12, 24]$. For the pre-training stage, the contrastive head uses a latent feature vector $v$ of size $512$. The VAE reconstruction head uses a kernel size of $3$ with 4 up-sampling stages. Our framework is implemented in PyTorch and MONAI on a single NVIDIA 2080Ti. The pre-training set is split into $80 \%$ for training and $20 \%$ for evaluation. We use the inpainting reconstruction L1 loss for our stopping criteria. For all semi-supervised strategies, we adopt $\lambda_{semi}= 1$ and used a stochastic gradient descent (SGD) with a learning rate of $lr = 1e^{-4}$. 

\section{Results}

\begin{table}[b]
\caption{Segmentation performance (Dice score) under varying amounts of labeled training data availability (Sec.~\ref{sec:training_variance}). Bold text indicates the best performer and  underlined text indicates the second best performer in each column.}
\resizebox{\columnwidth}{!}{%
\begin{tabular}{l|l|l|l}
\hline
\multirow{2}{*}{\textbf{Methods}}        & \multicolumn{3}{c}{Labeled training data size experiment} \\\cline{2-4}
                    
                                         & 3 labeled samples & 5 labeled samples & 10 labeled samples \\
                                         \hline
Fully supervised training on 7T only                    & 0.5211$\pm$0.0227    & 0.6235$\pm$0.0107     & 0.6872$\pm$0.0358      \\
Fully supervised training + Self-Supervised pre-train on 3T & 0.5329$\pm$0.0196    & 0.6138$\pm$0.0400     & 0.6995$\pm$0.0154      \\
Mean Teacher on 7T only                            & 0.5734$\pm$0.0210    & 0.6173$\pm$0.0371     & 0.7109$\pm$0.0249      \\
Cross Pseudo Supervision (CPS) on 7T only          & 0.6223$\pm$0.0321    & \underline{0.6885$\pm$0.0206 }    & 0.\underline{7541$\pm$0.0126}      \\
Mean Teacher + Self-Supervised pre-train on 3T & \underline{0.6358$\pm$0.0203}    & 0.6704$\pm$0.0184     & 0.7245$\pm$0.0193      \\
CPS + Self-Supervised pre-train  on 3T (SSL$^2$)      & \textbf{0.6565$\pm$0.0156}    & \textbf{0.7345$\pm$0.0307}     & \textbf{0.7915$\pm$0.0230}     \\
\hline
\end{tabular}
}
\label{tab:train_size}

\end{table}

\subsection{Labeled training data size experiment}
\label{sec:training_variance}
We first examine the performance of our model in the scenario of limited labeled data availability.
This is a fairly common scenario in practice: for example, often, a new MRI protocol may need  to be evaluated on a preliminary basis after the first few scans are acquired, before the entire dataset is collected.  In this scenario, very few annotated images are available to train, but we can generally assume that unlabeled images from previous similar studies likely exist. 

We split our labeled 7T in-house dataset (n=14) into 7 folds, with 12:2 train:test split in each fold. We evaluate three limited labeled dataset settings, using 3, 5 or 10 labeled images as training data. The remaining scans (12-3=9, 12-5=7, and 12-10=2, respectively) are treated as unlabeled data along with the unlabeled 7T in-house dataset (n=23).

\begin{figure}
    \centering
\includegraphics[width=0.8\columnwidth]{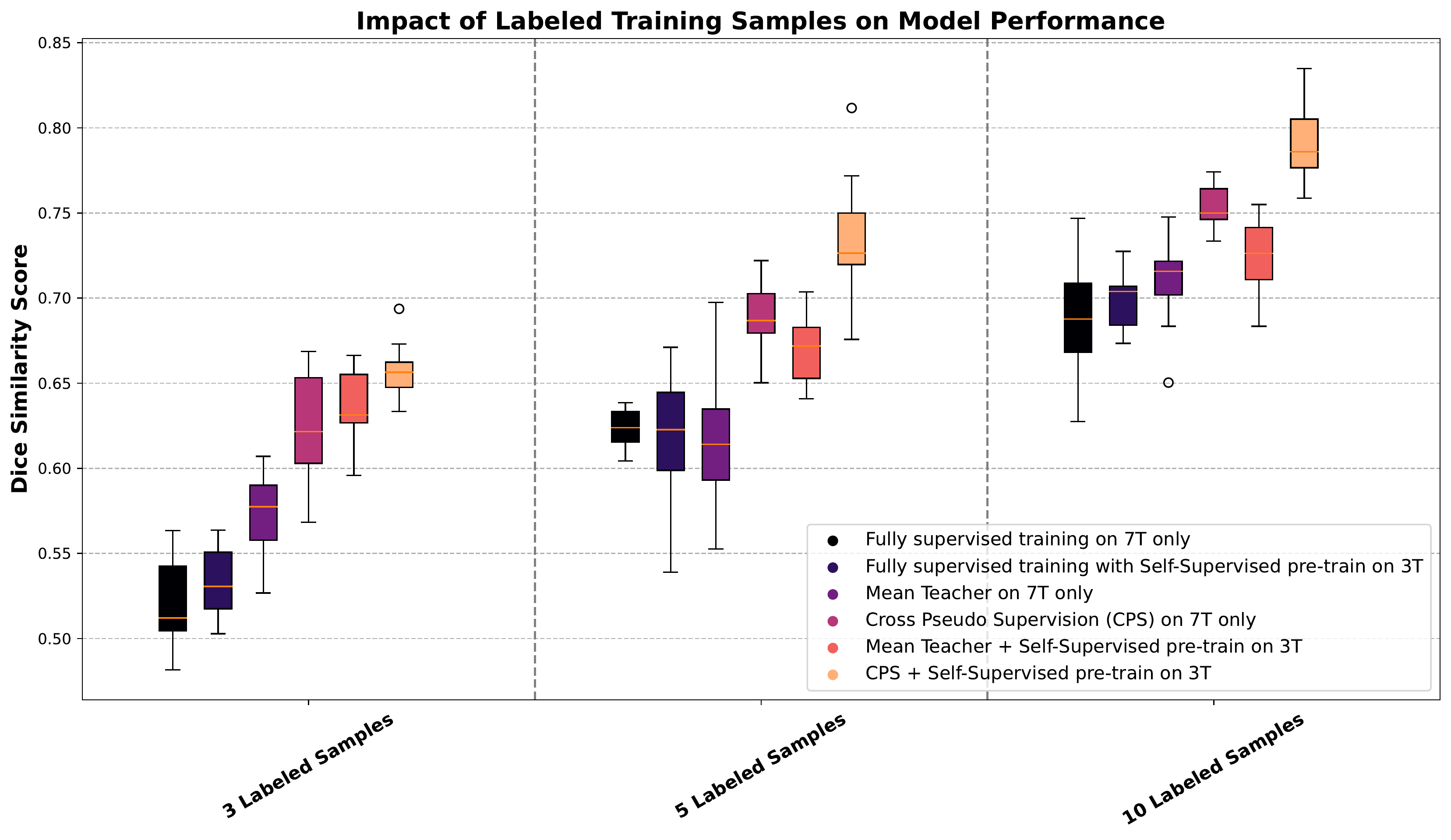}
    \caption{A box plot representation of the segmentation performance (Dice score) under varying amounts of labeled training data availability (Sec.~\ref{sec:training_variance}). The brightest box is our proposed framework and has the best performance; the darkest box is the baseline model of fully supervised training from scratch. }
    \label{fig:dice_samples}
\end{figure}

We present the 7-fold average Dice similarity coefficients in Table \ref{tab:train_size}. We observe that cross pseudo supervision (CPS) is again the best semi-supervised method and that it provides a dramatic performance increase (0.7915 vs.\ 0.6872 Dice) compared to the fully supervised model trained from scratch, i.e., on 7T data only. Even in the extreme scenario of only 3 training samples, our proposed framework can achieve comparable result to the fully supervised model with 10 training samples (0.6565 vs. 0.6872). We further observe that the 3T pre-training is beneficial to the 7T studies: the CPS with pre-training performs consistently better than CPS alone (e.g., 0.7541 vs. 0.7915 for 10 samples). These results are also presented in graphical format in Figure \ref{fig:dice_samples}.

\begin{table}[b]
\caption{Segmentation performance (Dice score) under varying amounts of sparsely labeled training data (Sec.~\ref{sec:sparse}). Bold text indicates the best performer and underlined text indicates the second best performer in each column.}
\label{tab:sparse}
\centering

\begin{tabular}{l|l|l|l|l}
\hline
\multirow{2}{*}{\textbf{Methods}}                 & \multicolumn{4}{c}{Sparse labeling experiment} \\ \cline{2-5}
                                         & 10 \%         & 20 \%         & 50 \%        & 100 \%       \\
                                         \hline
Fully supervised training on 7T only            & 0.5174        & 0.5732        & 0.6542       & 0.6971       \\
Fully supervised training + Self-Supervised pre-train on 3T   & 0.6233        & 0.6422        & 0.6855       & 0.7121       \\
Mean Teacher on 7T only                      & 0.5673        & 0.5884        & 0.6627       & 0.7047       \\
Cross Pseudo Supervision (CPS) on 7T only      & 0.6107        & 0.6342        & 0.7108       & \underline{0.7835}       \\
Mean Teacher + Self-Supervised pre-train on 3T & \underline{0.6411}        & \underline{0.6589}        & \underline{0.7232}       & 0.7624       \\
CPS + Self-Supervised pre-train  on 3T (SSL$^2$)         &\textbf{ 0.6523  }      & \textbf{0.6785}        & \textbf{0.7823}       & \textbf{0.8186 }     \\
\hline
\end{tabular}

\end{table}
\subsection{Sparse labeling experiment}
\label{sec:sparse}
We next evaluate our model in a sparse labeling scenario. It can be challenging to carefully and confidently label all the images in an MRI scan due to constraints on time and effort. A potential solution can be to only label a subset of the 2D slices in a given 3D MRI volume. This can also be useful when some 2D slices are difficult to annotate due to poor image quality or artifacts. To examine the use of such sparse annotations, we vary the percentage of annotated slices in a given training image. The remaining slices and data from the unlabeled set are pooled together and used as the new unlabeled cohort for the experiment. For example, in the $20\%$ setting, we randomly select 40 slices ($200\times0.20$) as labeled data and use the remaining 160 as unlabeled data. 

To allow a direct comparison to the experiments in Sec.\ \ref{sec:training_variance}, we randomly select 2 subjects for our test set and report the Dice score
in Table \ref{tab:sparse}. Similar to the training data size experiment, we note that CPS is the best performer. We observe that the self-supervised method contributes more than the semi-supervised method in this setting. We note that our proposed framework can outperform fully supervised training from scratch with $100 \%$ of the annotated data when using only $50 \%$ of annotations (0.7823 vs.\ 0.6971). 

\begin{figure}
    \centering
    \includegraphics[width=\columnwidth]{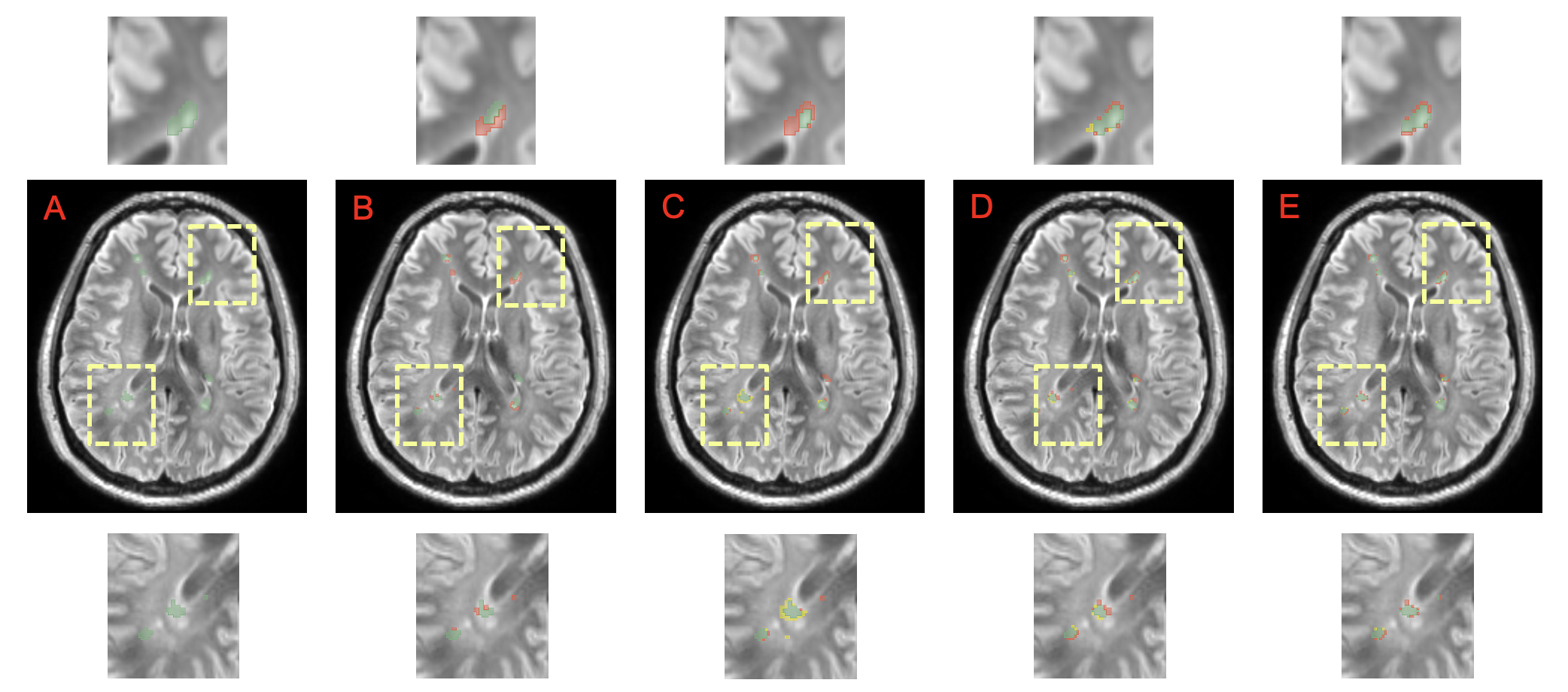}
    \begin{tabular}{P{0.17\columnwidth}P{0.17\columnwidth}P{0.17\columnwidth}P{0.17\columnwidth}P{0.17\columnwidth}}
        Ground Truth & Fully Supervised & Fully Supervised + & Mean Teacher + & CPS + \\
         & 7T only & Self-Supervised & Self-Supervised & Self-Supervised \\
        & & pre-train on 3T & pre-train on 3T & pre-train on 3T 
    \end{tabular}
    \caption{Qualitative  segmentation results in our 7T dataset with $100\%$ labeling and 12 labeled training samples. Segmentations are overlaid on the FLAIR image. Top and bottom rows show zoomed-in views of the yellow ROIs. Green denotes the true positives, red denotes false negatives, and yellow denotes false positives.  }
    \label{fig:qualitative}
\end{figure}

Finally, we present  qualitative results in Figure \ref{fig:qualitative}. Our proposed framework (panel E) yields the closest results to the manual ground truth annotations. The results of the 7T fully supervised MRI showed comparable performance in some areas, but had many false negatives (especially in the top zoom panel), which can be problematic for clinical diagnosis. The self-supervised learning approach improved the detection in some of these regions, but our proposed method combining semi-supervised and self-supervised learning resulted in the best performance. 

\section{Conclusion}
In this paper, we proposed a novel approach for achieving robust MS lesion segmentation in 7T brain MRI data by utilizing self-supervised training to embed information from publicly available 3T brain MRI data, in combination with semi-supervised techniques to leverage limited labeled 7T data. Our experimental results demonstrate the effectiveness of this approach, achieving higher accuracy with either a small number of training samples or sparsely annotated images.

 We make the pre-trained weights of our proposed approach publicly available to benefit future 7T brain MRI studies. In future work, we aim to investigate the generalizability of our proposed pre-trained encoder to other 7T MRI studies as well as its potential for use in downstream tasks beyond lesion segmentation.

\acknowledgments 

 This work was supported, in part, by NIH grant R01-NS094456 and National Multiple Sclerosis Society grant PP-1905-34001. Francesca Bagnato receives research support from Biogen Idec, the National Multiple Sclerosis Society
(RG-1901-33190) and the National Institutes of Health (1R01NS109114-01). Francesca Bagnato did not receive financial support for the research, authorship and publication of this article.

\bibliography{report} 
\bibliographystyle{spiebib} 

\end{document}